\documentclass[10pt,twocolumn,letterpaper]{article}

\usepackage[pagenumbers]{wacv}      

\usepackage[table]{xcolor}
\usepackage{arydshln} 
\usepackage{multicol}
\usepackage{multirow}
\usepackage{pifont}

\usepackage{graphicx} 

\graphicspath{{assets/}}

\newcommand{\cgr}[0]{\cellcolor{gray!20}}
\newcommand{\avgstd}[2]{#1 $\pm$ #2}
\newcommand{\Ours}[0]{\textcolor{blue}{\textbf{Ours}}}
\newcommand{\cmark}{{\color{OliveGreen}\ding{51}}}%
\newcommand{\xmark}{{\color{BrickRed}\ding{55}}}%

\definecolor{wacvblue}{rgb}{0.21,0.49,0.74}
\usepackage[pagebackref,breaklinks,colorlinks,allcolors=wacvblue]{hyperref}

\def\wacvPaperID{3116} 
\def\confName{WACV}
\def\confYear{2026}

\title{GFT: Graph Feature Tuning for Efficient Point Cloud Analysis}

\author{
    Manish Dhakal$^\dag$, Venkat R. Dasari$^\S$, Rajshekhar Sunderraman$^\dag$, Yi Ding$^\dag$\\
    $^\dag$Department of Computer Science, Georgia State University, Atlanta, GA\\
    $^\S$DEVCOM Army Research Laboratory, Aberdeen Proving Ground, MD\\
    {\tt\small mdhakal3@gsu.edu,venkateswara.r.dasari.civ@army.mil,rsunderraman@gsu.edu,yiding@gsu.edu}
}

\begin{document}
\maketitle
\begin{abstract}
Parameter-efficient fine-tuning (PEFT) significantly reduces computational and memory costs by updating only a small subset of the model's parameters, enabling faster adaptation to new tasks with minimal loss in performance.
Previous studies have introduced PEFTs tailored for point cloud data, as general approaches are suboptimal.
To further reduce the number of trainable parameters, we propose a point-cloud-specific PEFT, termed Graph Features Tuning (GFT), which learns a dynamic graph from initial tokenized inputs of the transformer using a lightweight graph convolution network and passes these graph features to deeper layers via skip connections and efficient cross-attention modules.
Extensive experiments on object classification and segmentation tasks show that GFT operates in the same domain, rivalling existing methods, while reducing the trainable parameters. Code is available at \url{https://github.com/manishdhakal/GFT}.
\end{abstract}    
\section{Introduction}
\label{sec:introduction}

Earlier methods of point cloud modeling include PointNet~\cite{qi2017pointnet}, PoinNet++~\cite{qi2017pointnet++}, and DGCNN~\cite{wang2019dynamic}. These methods utilize the global and local features from point clouds in $(x,y,z)$ coordinates space, yielding significant performance in classification and segmentation.
However, there has been evidence that these point cloud representation learning methods may not be optimal due to random initialization of weights~\cite{dong2023autoencoders,pang2022masked,yu2022point}.
One approach to fixing these issues is to perform self-supervised pretraining and full fine-tuning (FFT) for some downstream tasks.
This has been demonstrated to be effective in various works~\cite{liang2025pointmamba,liu2022masked,xie2020pointcontrast}. 

\begin{figure}[t]
    \centering
    \includegraphics[width=1.0\linewidth]{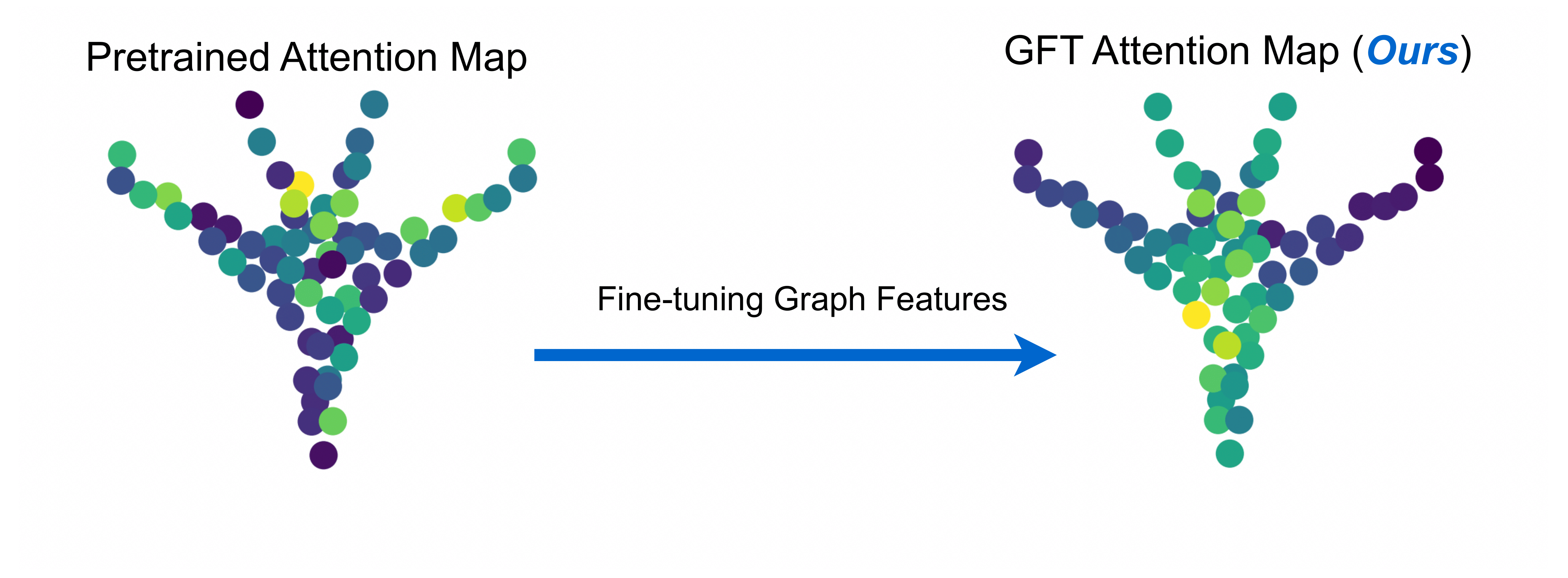}
    \caption{Attention maps (top-view) from the last layer with an \textbf{airplane} point cloud: Brighter points indicate patches contributing more to the global feature. The pretrained transformer (left) is random and uneven, while the GFT (right) shows more uniform regional pooling. Additional visualizations of different objects are included in the supplementary (\Cref{sec:suppl_vizualizations}).}
    \label{fig:attention_map}
\end{figure}
FFT of the pretrained models is expensive, as a separate copy of the fine-tuned model is required to be stored for every new additional tasks.
It demands more storage expense for horizontal scaling of the tasks and reduces the degree of parameters sharing.
To overcome this issue, various parameter-efficient fine-tuning (PEFT) methods---such as Adapters~\cite{houlsby2019parameter}, and VPT~\cite{jia2022visual},---have been developed for pretrained transformers~\cite{vaswani2017attention} networks.
These PEFTs have shown significant improvement in fine-tuning language and computer vision tasks.
However, \citet{zhou2024dynamic} exhibited the sub-optimality of these generic PEFTs to adapt downstream tasks for point cloud analysis, implying a need for PEFTs specific to point cloud~\cite{zha2023instance,zhou2024dynamic}.
IDPT~\cite{zha2023instance} and DAPT~\cite{zhou2024dynamic} rose as point-cloud-specific PEFTs to fine-tune downstream tasks.

Learning graph features within the tokens of transformers has been an intuitive approach for fine-tuning transformer~\cite{vaswani2017attention} models.
The motivations for incorporating graph features stems from the transformer architecture itself, where self-attention can be interpreted as a fully connected, dynamic graph with a complexity of $\mathcal{O}(n^2)$ over $n$ tokens, and the attention maps serve as edge weights.
For efficiency, we can make this dynamic graph update sub-quadratic or linear with $k\mathcal{O}(n)$ complexity ($k \ll n$) via KNN graphs.
Additionally, \Cref{fig:attention_map} demonstrates the effectiveness of fine-tuning graph features, as it rewrites the attention maps to capture discriminative local geometric structures—effectively learning where to focus. 
In contrast, the attention maps from the pretrained model appear random and scattered, lacking a clear sense of focus.

\begin{figure}[t]
    \centering
    \includegraphics[width=0.85\linewidth]{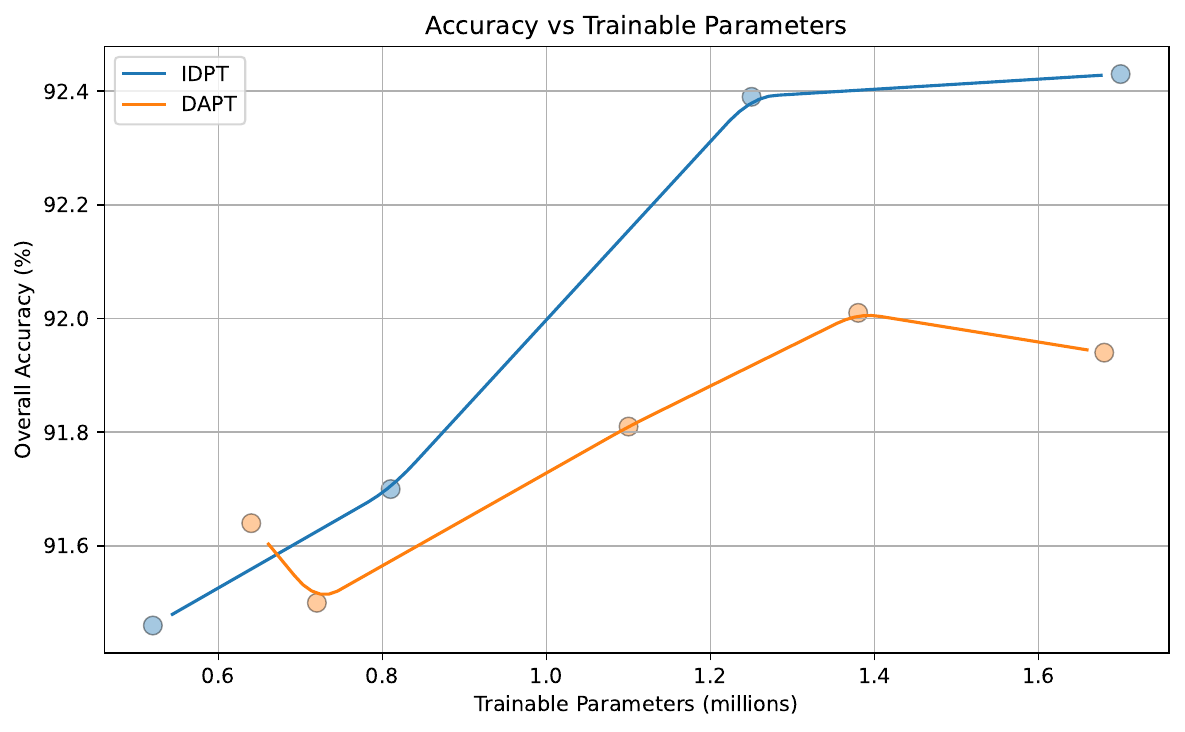}
    \caption{Within the parameter budget of $(0.6M, 1.7M)$, baselines (IDPT and DAPT) exhibited maximum drops of $0.97\%$ and $0.51\%$, respectively, when evaluated with OBG\_BJ dataset. Another ablation at \Cref{fig:params_vs_perf} compares our method with the baselines regarding performance-efficiency trade-off.}
    \label{fig:params_vs_perf_baslelines}
\end{figure}

IDPT~\cite{zha2023instance} employs a heavy graph features learner only in the model's final layer. 
However, this approach has a significant drawback: the model is relatively heavy and sensitive to parameter budget reduction (as shown in Figure~\ref{fig:params_vs_perf_baslelines}).
Our goal is to design a lightweight and point-cloud-specific PEFT that can learn graph features for fine-tuning while being parameter-budget-friendly.

\begin{figure*}[t]
    \centering
    \includegraphics[width=0.9\textwidth]{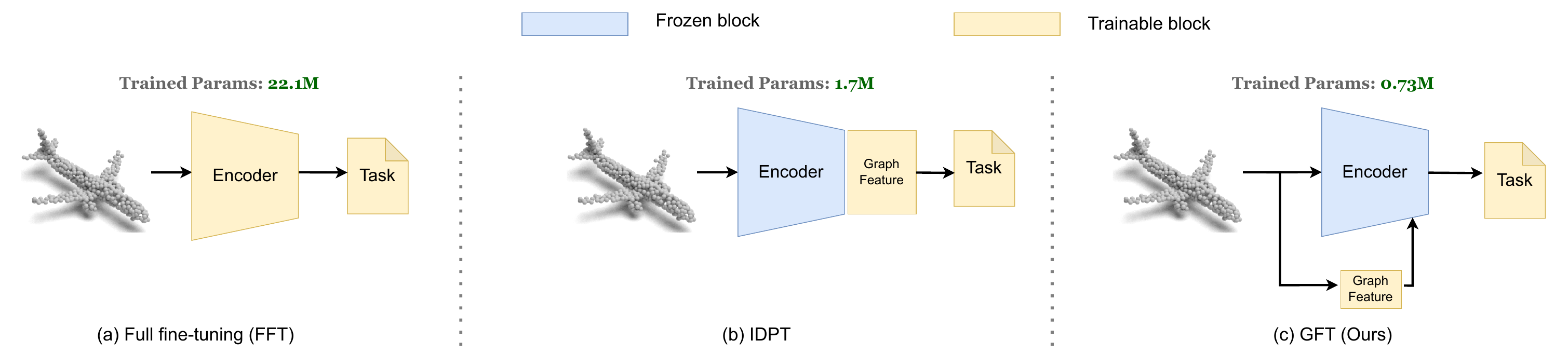}
    \caption{\textbf{Full fine-tuning (FFT) vs. IDPT~\cite{zha2023instance} vs. GFT. (classification task)} (a) The whole model is fine-tuned in an end-to-end manner, including the encoder and the task head. (b) IDPT freezes the encoders and trains a \textit{heavy} graph feature extractor at the last layer. (c) With \textit{light} blocks, GFT extracts graph features from the earliest tokens of transformers and selectively injects features into the encoder.}
    \label{fig:overall_idea}
\end{figure*}

With this study, we propose \textit{Graph Features Tuning (GFT)} that incorporates additional graph features from the transformer's input tokens to the deeper layers of the model, as shown in \Cref{fig:overall_idea}.
GFT introduces three modules that adapt downstream tasks: \textbf{(1) Task-specific prompt}~\cite{jia2022visual} module that prepends learnable continuous token to token embedding space for higher degree of freedom for the graph.
 \textbf{(2) EdgeConv}~\cite{wang2019dynamic} module has multiple layers of graph convolutional network (GCN) that embed locality information. 
\textbf{(3) Cross-attention interaction} module injects the graph features into the encoder. 
All of these modules are extremely small, updating merely 0.73 million parameters to fine-tune a pretrained transformer for classification.

In summary, we highlight our major contributions as:
\begin{itemize}
    \item We propose Graph Feature Tuning (GFT), a novel fine-tuning method specifically designed for point cloud data, which achieves state-of-the-art parameter efficiency to the best of our knowledge.
    \item Through extensive experiments and ablation studies, we systematically analyze the performance-efficiency trade-off, demonstrating that GFT maintains competitive accuracy while significantly reducing the number of trainable parameters compared to existing benchmarks.
\end{itemize}

\section{Related Work}
\label{sec:relatedwork}

\subsection{Representation Learning of Point Clouds}

Earlier self-supervised models used convolution-like operators~\cite{wang2019dynamic,wu2019pointconv} to pass messages between points and their neighbours~\cite{zhang2022pointdae,wang2021unsupervised,zhang2021self} for representation learning.
With the rise of Transformers~\cite{vaswani2017attention}, Point Transformers~\cite{zhao2021point} have been devised to encode a group of points as a single token, and multiple sequential tokens were used as inputs for stacked self-attention networks~\cite{pang2022masked,zhang2022point,dong2023autoencoders,yu2022point,qi2023contrast}.
Following the principles of masked language modeling, most of the self-supervised point cloud models learn from masked point modeling that randomly masks points from the cloud and enforces the models to predict the missing signal.
Point-BERT~\cite{yu2022point} learns to generate discrete point tokens containing meaningful information about the missing signals by using discrete Variational AutoEncoder (dVAE).
Point-MAE~\cite{pang2022masked}, PointM2AE~\cite{zhang2022point}, and ReCon~\cite{qi2023contrast} have shown significant improvement in the pertaining process by reconstructing the missing information.
ACT~\cite{dong2023autoencoders} learns the representation with self-supervision by employing cross-modal transfer of weights from a 2D image or language encoder.
In our work, we use the pretrained models with architectures based on point transformers trained with different reconstruction methods and a variety of pertaining recipes.

\subsection{Parameter Efficient Fine-Tuning for Point-Clouds} 
Parameter-efficient fine-tunings (PEFTs) modulate the representation of the pretrained models with addition of fewer learnable parameters for the downstream tasks.
Adapters update pretrained models by training smaller blocks that runs in parallel to pretrained blocks~\cite{houlsby2019parameter, rebuffi2017learning}, effectively fine-tuning vision and language tasks~\cite{houlsby2019parameter, wu2025slora, dhakal2024vlsm, hu2022lora}. 
Prompt tuning, on the other hand, introduces learnable continuous tokens into the transformer's embedding space while keeping the pretrained model frozen~\cite{jia2022visual, zhou2022learning, shin2020autoprompt, gao2021making, adhikari2024tunevlseg}.  
Although these PEFT methods have demonstrated success in vision and language domains, they fall short in point cloud analysis~\cite{zhou2024dynamic, zha2023instance}. 
To address this, two PEFT approaches have been specifically designed for point clouds: IDPT~\cite{zha2023instance} and DAPT~\cite{zhou2024dynamic}. 
IDPT learns prompts from the last layer by generating a graph between tokens, while DAPT fine-tunes adapters by generating a dynamic scale for each token.  
In this work, we propose a PEFT that further optimizes efficiency for point cloud tasks.

\subsection{Message Passing in Point Cloud}
Message passing from neighboring data to generate local features is straightforward for text and 2D images because they inherently possess structural information. In these domains, convolutional operators can efficiently aggregate information from neighboring elements within the feed-forward network~\cite{he2016deep,conneau2016very,kalchbrenner2014convolutional}. However, point clouds lack such structural information, as they are represented as an unordered set of $xyz$ coordinates. To enable convolution-like message passing, a graph-based structure is required to define relationships between points, typically using K-nearest neighbors (KNN)~\cite{qi2017pointnet++,wang2019dynamic}. In this work, we utilize EdgeConv~\cite{wang2019dynamic} to encode locality features from the token space of the transformer, allowing for effective locality feature extraction.

\section{Methodology}

\begin{figure*}[t]
    \centering
    \includegraphics[width=0.95\textwidth]{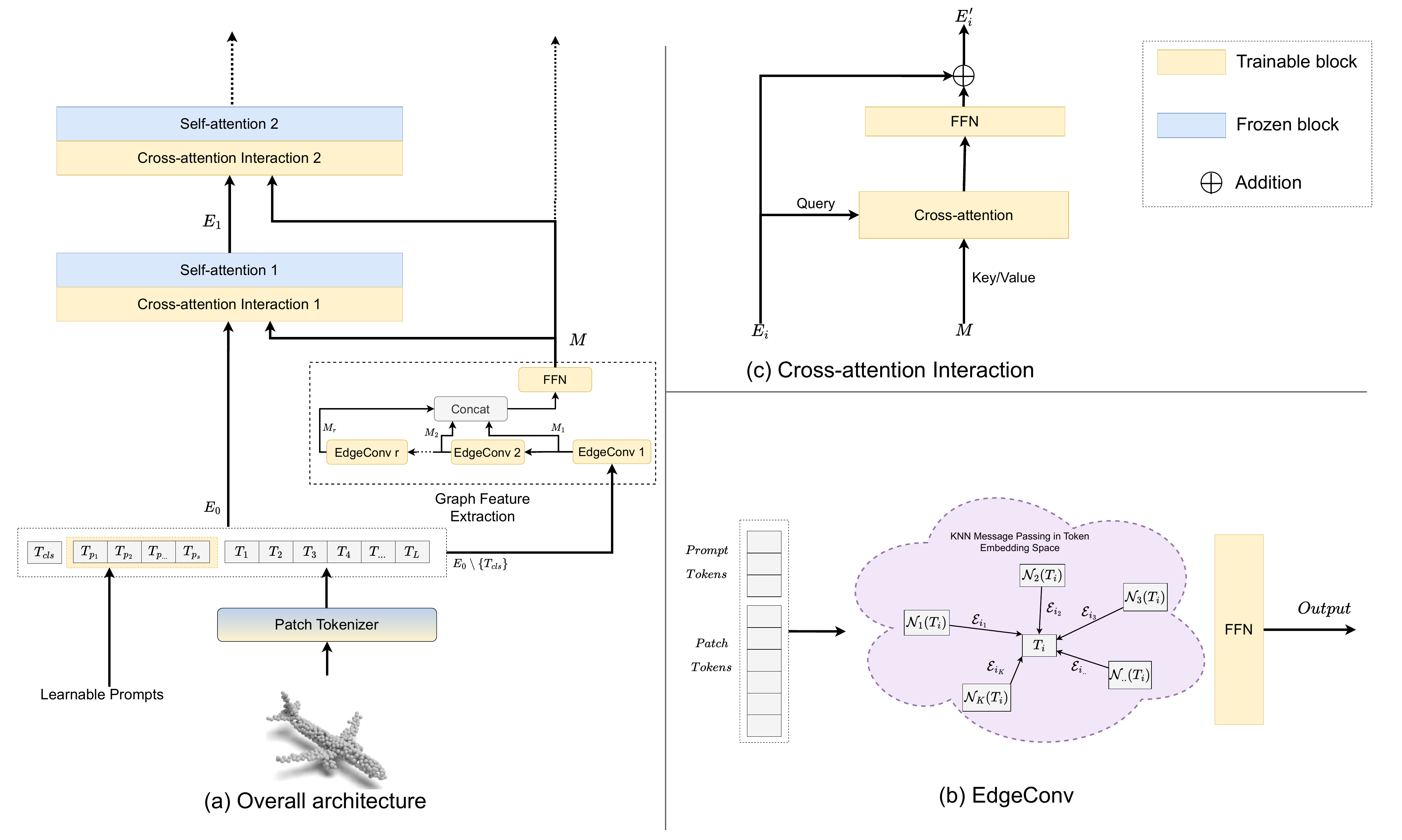}
    \caption{\textbf{Overall architecture of GFT.} (a) GFT is the composition of learnable prompts, graph feature extraction, and cross-attention interaction. (b) EdgeConv generates graph features from K-nearest neighbouring tokens, where the edge feature is $\mathcal{E}_i=\mathcal{N}(T_i)-T_i$. (c) Cross-attention interaction blocks use the encoder features $E_i$ as query and the graph features $M$ as key/value for the attention module.}
    \label{fig:architecture}
\end{figure*}

\subsection{Preliminary of Point Transformer}
Point Transformer~\cite{zhao2021point} encodes 3D points $x \in \mathbb{R}^{N\times3}$ from $(x,y,z)$ space to $D$-dimensional latent space.
It contains two major components: a point tokenizer that divides the entire $N$ points into $L$ tokens and $F$ (usually $F=12$) transformer encoder layers with self-attention mechanisms.

First, the point tokenizer selects $L$ token centroids with the farthest point sampling (FPS) method from the entire set of points.
Then, each center samples its K-nearest neighbours to embed the locality information, forming the $L$ tokens as $x^\prime \in \mathbb{R}^{L\times k \times 3}.$
Each of the token is projected to $D$ dimension space to form a 1-dimensional grouped point representation as a matrix of shape $\mathbb{R}^{L \times D}$. 
A classification token, $T_{cls} \in \mathbb{R}^{1\times D}$, is prepended to the tokens which give the initial embedding matrix: 
\begin{equation} \label{eq:initial_embedding}
E_0 = [T_{cls}; [T_1, T_2, T_3,...,T_L]],
\end{equation}
where $E_0 \in \mathbb{R}^{(1+L) \times D}$. This embedding serves as the input to the first self-attention layer. 
The output from each self-attention layer is passed as input to the next layer as:
\begin{equation}
    E_i = SelfAttention(E_{i-1}), i \in \{1,2,....,F\}
\end{equation}

\subsection{Graph Feature Tuning (GFT)}
\label{sec:gft}
Illustrated in \Cref{fig:architecture}, GFT has three major components for fine-tuning. 
First, we prepend learnable continuous prompts to the sequential tokens so the graphs have a higher degree of freedom.
Second, the graph features extractor, EdgeConv, extracts graph representations from the token embedding space.
It takes input from the point tokenizer block and generates multi-layer graph features.
We unlock the first layer of tokenizer to receive the parameter updates for unrestricted representation of the tokens.
Third, we use cross-attention modules to inject features into the encoder.
The features injection takes place sparsely to make it lightweight.

\subsubsection{Task-specific Learnable Prompt}
Learnable prompt tuning is one of the efficient fine-tuning methods used in the 2D-vision and language domains to learn task-specific features\cite{jia2022visual,gao2021making}.
This extremely parameter-efficient method injects learnable continuous prompts into the token embedding space of the transformer while the entire encoder remains frozen.
Similarly, we inject these learnable prompts to the initial embeddings $E_0$ of \Cref{eq:initial_embedding} between $T_{cls}$ and other tokens as:
\begin{align}
    E_0 & := Inject(E_0, P), \\
       {} & := [T_{cls}; P; [T_1,T_2,...,T_L]],\\
       {} & := [T_{cls}; [T_{P_1},T_{P_2},...,T_{P_s}]; [T_1,T_2,...,T_L]], \label{eq:prompt_injection}
\end{align}

where, $P \in \mathbb{R}^{s \times D}$ denotes learnable prompt and updated tokens representation $E_0 \in \mathbb{R}^{(1+s+L) \times D}$.

\subsubsection{EdgeConv Feature Extraction}
EdgeConv~\cite{wang2019dynamic} extracts features from the initial embedding matrix $E_0$ (excluding the $T_{cls}$ token) of the transformer pipeline.
It is a component of multi-layered graph convolution functions for message passing that encodes features from the immediate neighbours.
However, $E_0$ does not contain the intrinsic neighbourhood information of a token.
Thus, we define the locality information of $i^{th}$ token $T_i \in \mathbb{R}^{1 \times D}$ by grouping its K-nearest neighbours $\mathcal{N}(T_i) \in \mathbb{R}^{K \times D}$.
Distance from the neighbours to $T_i$ is attributed as edge features for KNN graphs, as shown in \Cref{fig:architecture} (b):
\begin{equation}
    \mathcal{E}_i = \mathcal{N}(T_i) - T_i,
\end{equation}
where $\mathcal{E}_i \in \mathbb{R}^{K \times D}$.
We concatenate the $\mathcal{E}_i$ with the central token $T_i$ in the feature dimension, which gives us the neighbourhood information of $T_i$ as $\mathcal{\bar{N}}(T_i) \in \mathbb{R}^{K \times 2D}$.
Projecting the neighbourhood information to a latent space of $d_1$ and pooling across $K$ dimension, we get a new neighbourhood embedding of $T_i$ as $\mathbb{R}^{1 \times d_1}$. 
We apply this for all $(s+L)$ tokens to get the learned graph feature as $M_1 \in \mathbb{R}^{(s+L) \times d_1}$.
This process is continued for multiple layers of latent spaces, creating a pyramid of graph features $[M_1,M_2,....M_r]$, where $r$ is the depth of the feature pyramid. These pyramid features are concatenated and fed to a feed-forward network (FFN) as:

\begin{equation}
    M = FFN(Concat[M_1, M_2,....M_r]))
\end{equation}
where $M \in \mathbb{R}^{(s+L) \times d} $ gives the final output of the EdgeConv network.
For our experimentation, we work with $r=4$ and each feature dimension of $64$ for the pyramid.

\subsubsection{Cross-attention Interactions}
Features from EdgeConv interact with the pretrained Point-Transformer to update the transformer's features.
Each interaction includes a cross-attention module~\cite{vaswani2017attention} to fine-tune the pretrained features.
To make the interaction lightweight, we implement sparse interaction to self-attention layers of point transformer.
For instance, our implementation of $F=12$ layered encoders is subjected to $4$ interactions at $i \in \{1^{st},4^{th},7^{th},10^{th}\}$ self-attention layers.

At the input of $i^{th}$ transformer encoder, the transformer embeddings $E_{i-1}$ from the previous layer are updated by cross-attention modules operating with EdgeConv pyramid features $M$, where $E_{i-1}$ and $M$ are queries and keys features:
    \begin{align}
        E^\prime_{i-1} & = XAttn(norm(E_{i-1}), norm(M)),
    \end{align}
where $norm(\cdot)$ is layer normalization and $XAttn(a, b)$ is a cross-attention block, $a$ denotes query, and $b$ denotes key/value.  The tensor shape of the embedding remains unchanged for use in the next encoder block.

\begin{table*}[t]
    \centering
    \resizebox{\linewidth}{!}{%
        \begin{tabular}{llcccccc}
            
            \hline
              \multirow{2}{*}{\textbf{Method}} & \multirow{2}{*}{\textbf{Venue}}  & \multirow{2}{*}{\textbf{\#T-Params (M) $\downarrow$}} & \multirow{2}{*}{\textbf{FLOPs(G)}} & \multicolumn{3}{c}{ \textbf{ScanObjectNN}}   & \multirow{2}{*}{\textbf{ModelNet40}} \\
             \cmidrule{5-7}
             &  & & & OB\_JBG  & OBJ\_ONLY  & PB\_T50\_RS  &  \\
             \midrule
             \multicolumn{8}{c}{\textit{Training with random weights initialization}}\\
             \midrule
             PointNet~\cite{qi2017pointnet} & CVPR'17 & 3.5 & 0.5 & 73.3 & 79.2 & 68.0 & 89.2\\
             PointNet++~\cite{qi2017pointnet++} & NeurIPS'17 & 1.5 & 1.7 & 82.3 & 84.3 & 77.9 & 90.7 \\
             DGCNN~\cite{wang2019dynamic} & TOG'19 & 1.8 & 2.4 & 82.8 & 86.2 & 78.1 & 92.9 \\
             MVTN~\cite{hamdi2021mvtn} & ICCV'21 & 11.2 & 43.7 & - & - & 82.8 & 93.8 \\ 
             PointNeXt~\cite{qian2022pointnext} & NeurIPS'22 & 1.4 & 1.6 & - & - & 87.7 & 94.0 \\
             PointMLP~\cite{ma2022rethinking} & ICLR'22 & 13.2 & 31.4 & - & - & 85.4 & 94.5 \\
             RepSurf-U~\cite{ran2022surface} & CVPR'22 & 1.5 & 0.8 & - & - & 84.3 & 94.4 \\
             ADS~\cite{hong2023attention} & ICCV'23 & - & - & - & - & 87.5 & 95.1 \\
             \midrule
             \multicolumn{8}{c}{\textit{Full fine-tuning (FFT) of pretrained models}}\\
             \midrule
             OcCo~\cite{wang2021unsupervised} & ICCV'21 & 22.1 & 4.8 & 84.85 & 85.54 & 78.79 & 92.1 \\
             MaskPoint~\cite{liu2022masked}& ECCV'22 & 22.1 & - & 89.70 & 89.30 & 84.60 & 93.8 \\
             Point-M2AE~\cite{zhang2022point}& NeurIPS'22 & 15.3 & 3.6 & 91.22 & 88.81 & 86.43 & 94.0 \\
             ReCon\cite{qi2023contrast} & ICML'23 & 43.6 & 5.3 & 94.15 & 93.12 & 89.73 & 93.9 \\
             \midrule
             Point-MAE~\cite{pang2022masked} & ECCV'22 & 22.1 & 4.8 & 90.02 & 88.29 & 85.18 & 93.8 \\
             Point-BERT~\cite{yu2022point} & CVPR'22 & 22.1 & 4.8 & 87.43 & 88.12 & 83.07 & 93.2 \\
             ACT~\cite{dong2023autoencoders}& ICLR'23 & 22.1 & 4.8 & 93.29 & 91.91 & 88.21 & 93.7 \\
             
             \midrule
             \multicolumn{8}{c}{\textit{Parameter efficient fine-tuning (PEFT) with different backbones}}\\
             \midrule
             \multicolumn{8}{c}{Point-MAE backbone} \\
             \hdashline
              IDPT~\cite{zha2023instance} & ICCV'23 & 1.70 (7.69\%) & 7.2 & \avgstd{92.43}{0.27}  & \avgstd{91.43}{0.28} & \avgstd{\textbf{88.27}}{0.23} & \avgstd{93.00}{0.18} \\
             DAPT~\cite{zhou2024dynamic} & CVPR'24 & 1.10 (4.97\%) & 5.0 & \avgstd{91.81}{0.36}  & \avgstd{91.36}{0.76} & \avgstd{87.56}{0.64} & \avgstd{92.88}{0.21} \\
             \cgr GFT (\Ours) & \cgr - & \cgr \textbf{0.73 (3.26\%)} & \cgr 6.4 & \cgr \avgstd{\textbf{92.56}}{0.58} &  \cgr \avgstd{\textbf{91.46}}{0.66}  & \cgr \avgstd{86.78}{0.32} & \cgr \avgstd{\textbf{93.19}}{0.19} \\
            \midrule
            \multicolumn{8}{c}{Point-BERT backbone} \\
            \hdashline
             IDPT~\cite{zha2023instance} & ICCV'23 & 1.70 (7.69\%) & 7.2 & \avgstd{91.98}{0.76}  & \avgstd{91.67}{0.78} & \avgstd{86.91}{0.78} & \avgstd{92.67}{0.17} \\
             DAPT~\cite{zhou2024dynamic} & CVPR'24 & 1.10 (4.97\%) & 5.0 &\avgstd{\textbf{93.05}}{0.59}  & \avgstd{\textbf{91.67}}{0.50} & \avgstd{\textbf{88.63}}{0.19} & \avgstd{92.49}{0.08} \\
             \cgr GFT (\Ours) & \cgr - \cgr & \cgr \textbf{0.73 (3.26\%)}  & \cgr 6.4 & \cgr \avgstd{92.36}{0.36} & \cgr \avgstd{91.22}{0.32} & \cgr \avgstd{87.37}{0.34} & \cgr \avgstd{\textbf{93.01}}{0.18} \\
            \midrule
             \multicolumn{8}{c}{ACT backbone} \\
            \hdashline
             IDPT~\cite{zha2023instance} & ICCV'23 & 1.70 (7.69\%) & 7.2 & \avgstd{91.60}{0.39}  & \avgstd{\textbf{91.60}}{0.37} & \avgstd{\textbf{87.15}}{0.39} & \avgstd{\textbf{93.07}}{0.20} \\
             DAPT~\cite{zhou2024dynamic} & CVPR'24 & 1.10 (4.97\%) & 5.0 &\avgstd{91.15}{0.20}  & \avgstd{91.15}{0.52} & \avgstd{86.97}{0.45} & \avgstd{92.63}{0.21} \\
             \cgr GFT (\Ours) & \cgr - & \cgr \textbf{0.73 (3.26\%)} & \cgr 6.4 & \cgr \avgstd{\textbf{92.39}}{0.48} & \cgr \avgstd{91.26}{0.28} & \cgr \avgstd{86.46}{0.43} & \cgr \avgstd{93.07}{0.21} \\
             \bottomrule
        \end{tabular}%
    }
    \caption{\textbf{Overall accuracy (OA\% $\uparrow$) of classification on three variants of the ScanObjectNN~\cite{uy2019revisiting} and the ModelNet40~\cite{wu20153d}}. \#T-params and FLOPS (G) indicate the number of trainable parameters (in millions) and inference floating point operations, respectively. ScanObjectNN used the strong data augmentation mentioned by ACT \cite{dong2023autoencoders}.}
    \label{tab:main_results}
\end{table*}

\section{Experimental Setup}
\label{sec:experimental_setup}
\subsection{Baseline PEFTs}
To cover a wider spectrum of pretraining strategies for Point Transformer, we include three backbones:  Point-BERT~\cite{yu2022point}, Point-MAE~\cite{pang2022masked}, and ACT~\cite{dong2023autoencoders}.
These foundation models were pretrained with ShapeNetCore from ShapeNet~\cite{chang2015shapenet}, containing $\sim$51K CAD models of 55 common object categories.
We include the results from two point-cloud-specific PEFTs, IDPT~\cite{zha2023instance} and DAPT~\cite{zhou2024dynamic}, to compare with our method.

\subsection{Classification}
For fair and robust comparison, we reproduce the results from IDPT~\cite{zha2023instance} and DAPT~\cite{zhou2024dynamic} under the same experimental setting as ours.
In \Cref{tab:main_results}, we report the average overall accuracy (OA\%) of models with five consecutive seeds within $[0,4]$.
We design the task head to take inputs from CLS token $T_{cls}$ and pooled prompt and patch tokens from the last layer of the transformer.
Two datasets are used for evaluations: ScanObjectNN for real-world object classification and ModelNet-40 for synthetic object classification. 

\noindent\textbf{ScanObjectNN}~\cite{uy2019revisiting} is a real-world point cloud dataset containing scanned objects from indoor scenes. 
It consists of approximately 15K object scans with background clutter and occlusions, categorized into 15 object types. 
We use three dataset splits: OBJ\_BG, OBJ\_ONLY, and PB\_T50\_RS. For preprocessing, we sample 2K points per instance and apply a simple rotation, inspired by ACT~\cite{dong2023autoencoders}. 

\noindent \textbf{ModelNet-40}~\cite{wu20153d} consists of 12,311 CAD models of synthetic objects across 40 different categories. The point clouds in these models are clean, noise-free, and uniformly distributed with independent and identical distributions. Following the standard approach, we split the dataset into 9,843 instances for training and 2,468 for testing. During training, we sample 1K points from the distribution and apply random scaling and translation. 

\subsection{Segmentation}
For segmentation, we design a transformer decoder with skip connections from the 3rd, 6th, 9th, and 11th layers of the pretrained model.
Architecture used for the decoder of the segmentation is provide in the supplementary (\Cref{sec:suppl_seg_head}).
In line with the baseline papers IDPT~\cite{zha2023instance} and DAPT~\cite{zhou2024dynamic}, we report the best performance across ten consecutive seeds, $[0,9]$, as reproducing the baselines to compute mean metrics was computationally demanding.
The evaluation dataset for the segmentation task is ShapeNetPart.

\noindent \textbf{ShapeNetPart}~\cite{yi2016scalable} is a part segmentation dataset containing 50 different parts across 15 object categories. Following the standard benchmarking protocol~\cite{zhou2024dynamic, zha2023instance}, we divide the dataset into training and test splits, consisting of 14K and 2.9K scans, respectively. Each data instance is represented by 2K sampled points. 

\subsection{Implementation Details}
We employed identical experimental settings to the PEFT methods for each baseline to ensure a fair comparison.
All experiments are conducted on a single GeForce RTX 4090.
Other hyperparameters regarding the fine-tuning tasks are mentioned in the supplementary (\Cref{sec:supp_train_details}).
\section{Results}
\label{sec:results}

\paragraph{Real-World Object Classification.}
The baseline foundational models favor clean and uniform point cloud data, similar to the pretraining dataset ShapeNet~\cite{chang2015shapenet}. However, real-world point clouds have a different distribution due to noise, missing data, and non-uniform point density.
Thus, we benchmark with ScanObjectNN~\cite{uy2019revisiting}.

As shown in \Cref{tab:main_results}, our method achieves an average performance gain of 4.12\% over the FFT baseline of Point-BERT and 2.47\% over Point-MAE. However, compared to the ACT baseline, there is an average performance drop of 0.5\%. This could be attributed to ACT’s multi-stage pretraining~\cite{dong2023autoencoders}, which, while effective, is computationally inefficient.  

Notably, our method achieves competitive results with only 0.73\% of the original parameter size. Compared to other PEFT methods, it outperforms most DAPT counterparts. While it does not surpass IDPT, it remains competitive. However, IDPT trains 1.7M parameters, more than twice the size of our model.

\paragraph{Synthetic Objects Classification}
ModelNet40~\cite{wu20153d} contains synthetic models of objects.
As illustrated in \Cref{tab:main_results}, our method does not exceed the FFT baselines of the pretrained backbones but remains competitive, with only a 0.5\% average accuracy deficit. This finding of FFT can be attributed to the high similarity between the pretraining and fine-tuning datasets, as both consist of synthetic CAD models with a clean and uniform point distribution. Notably, our method outperforms other PEFT approaches across all three pretrained backbones.

\paragraph{Few-shot Classification}

\begin{table}[t]
  \centering
  \resizebox{\linewidth}{!}{
    \begin{tabular}{lcccc}
    \toprule
    \textbf{Method} & \multicolumn{2}{c}{5-way} & \multicolumn{2}{c}{10-way} \\
\cmidrule{2-5}          & 10-shot & 20-shot & 10-shot & 20-shot \\
    \midrule
    \multicolumn{5}{c}{\textit{Full fine-tuning (FFT) of pretrained models}} \\
    \midrule
    DGCNN-OcCo~\cite{wang2021unsupervised}& \avgstd{90.6}{2.8} & \avgstd{92.5}{1.9} & \avgstd{82.9}{1.3} & \avgstd{86.5}{2.2} \\
    Transformer-OcCo~\cite{yu2022point} & \avgstd{94.0}{3.6} & \avgstd{95.9}{2.3} & \avgstd{89.4}{5.1} & \avgstd{92.4}{4.6} \\
    MaskPoint~\cite{liu2022masked}  & \avgstd{95.0}{3.7} & \avgstd{97.2}{1.7} & \avgstd{91.4}{4.0} & \avgstd{93.4}{3.5} \\
    EPCL\cite{huang2022frozen}  & \avgstd{95.1}{2.7} & \avgstd{97.3}{1.6} & \avgstd{91.1}{4.2} & \avgstd{93.5}{3.8} \\
    Point-M2AE~\cite{zhang2022point} & \avgstd{96.8}{1.8} & \avgstd{98.3}{1.4} & \avgstd{92.3}{4.5} & \avgstd{95.0}{3.0} \\
    \midrule
    Point-MAE~\cite{pang2022masked} & \avgstd{96.3}{2.5} & \avgstd{97.8}{1.8} & \avgstd{92.6}{4.1} & \avgstd{95.0}{3.0} \\
    Point-BERT~\cite{yu2022point} & \avgstd{94.6}{3.1} & \avgstd{96.3}{2.7} & \avgstd{91.0}{5.4} & \avgstd{92.7}{5.1} \\
    ACT~\cite{dong2023autoencoders} & \avgstd{96.8}{2.3} & \avgstd{98.0}{1.4} & \avgstd{93.3}{4.0} & \avgstd{95.6}{2.8} \\
    \midrule
    \multicolumn{5}{c}{\textit{Parameter efficient fine-tuning (PEFT) with different backbones}} \\
    \midrule
    \multicolumn{5}{c}{Point-MAE backbone} \\
    \hdashline
    IDPT~\cite{zha2023instance} & \avgstd{\textbf{97.3}}{2.1} & \avgstd{97.9}{1.1} & \avgstd{92.8}{4.1} & \avgstd{95.4}{2.9} \\
    DAPT~\cite{zhou2024dynamic} & \avgstd{96.8}{1.8} & \avgstd{98.0}{1.0} & \avgstd{\textbf{93.0}}{3.5} & \avgstd{\textbf{95.5}}{3.2} \\
    \cgr GFT (\Ours) & \cgr \avgstd{96.3}{3.1} & \cgr \avgstd{\textbf{98.3}}{1.5} & \cgr \avgstd{92.1}{5.3} & \cgr \avgstd{95.1}{2.9} \\
    \midrule
    \multicolumn{5}{c}{Point-BERT backbone} \\
    \hdashline
    IDPT~\cite{zha2023instance} & \avgstd{\textbf{96.0}}{1.7} & \avgstd{97.2}{2.6} & \avgstd{91.9}{4.4} & \avgstd{93.6}{3.5} \\
    DAPT~\cite{zhou2024dynamic} & \avgstd{95.8}{2.1} & \avgstd{97.3}{1.3} & \avgstd{\textbf{92.2}}{4.3} & \avgstd{94.2}{3.4} \\
    \cgr GFT (\Ours) & \cgr \avgstd{96.0}{2.5} & \cgr \avgstd{\textbf{98.1}}{1.3} & \cgr \avgstd{92.2}{4.5} & \cgr \avgstd{\textbf{94.5}}{3.7} \\
    \midrule
    \multicolumn{5}{c}{ACT backbone} \\
    \hdashline
    IDPT~\cite{zha2023instance} & \avgstd{\textbf{96.4}}{2.5} & \avgstd{98.3}{1.6} & \avgstd{\textbf{92.5}}{4.5} & \avgstd{\textbf{95.6}}{2.9} \\
    DAPT~\cite{zhou2024dynamic} & \avgstd{96.0}{3.0} & \avgstd{97.4}{2.2} & \avgstd{90.5}{4.9} & \avgstd{94.2}{3.6} \\
    \cgr GFT (\Ours) & \cgr \avgstd{96.0}{2.3} & \cgr \avgstd{\textbf{98.3}}{1.5} & \cgr \avgstd{92.1}{5.0} & \cgr \avgstd{95.0}{3.4} \\
    \bottomrule
    \end{tabular}%
  }
  \caption{\textbf{Few-shot learning on ModelNet40~\cite{wu20153d}.} We report the average overall accuracy (\%) with the standard deviation (\%) of 10 independent experiments.}
  \label{tab:few_shot}%
\end{table}%

Following previous works~\cite{zhou2024dynamic, zha2023instance, dong2023autoencoders, pang2022masked, yu2022point}, we evaluate the pretrained model on \textit{N-way-K-shot} few-shot classification. In this setting, \( N \in \{5,10\} \) represents the number of object categories, while \( K \in \{10,20\} \) denotes the number of training instances per category. For example, a \textit{5-way-10-shot} setup consists of five classes, each with ten training samples, totaling 50 training instances. We use a modified version of the ModelNet-40~\cite{wu20153d} dataset for this task, where classes and training instances are sampled from its 40 categories.  

As reported in \Cref{tab:few_shot}, our method outperforms most existing FFT and PEFT approaches. While it may not surpass all PEFT baselines (IDPT and DAPT), across every \textit{N-way-K-shot} scenario, it achieves significant reductions of 57\% and 34\% in parameter space, respectively. Even in cases where performance lags slightly behind, the loss is negligible compared to the substantial improvement in parameter efficiency.

\paragraph{Part Segmentation}
\begin{table}[t]
  \centering
  \resizebox{\linewidth}{!}{
    \begin{tabular}{lccc}
    \toprule
    \textbf{Method} & \#T-Params (M) & Cls. mIoU (\%) & Inst. mIoU (\%) \\

    \midrule
    \multicolumn{4}{c}{\textit{Training with random weight initialization}} \\
    \midrule
    PointNet~\cite{qi2017pointnet} & - & 80.39 & 83.7 \\
    PointNet++~\cite{qi2017pointnet++} & - & 81.85 & 85.1 \\
    DGCNN~\cite{wang2019dynamic} & - & 82.33 & 85.2 \\
    APES~\cite{wu2023attention} & - & 83.67 & 85.8 \\

    \midrule
    \multicolumn{4}{c}{\textit{Full fine-tuning (FFT) of pretrained models}} \\
    \midrule
    OcCo~\cite{wang2021unsupervised} & 27.09 & 83.42 & 85.1 \\
    MaskPoint~\cite{liu2022masked} & - & 85.60 & 86.0 \\
    ACT~\cite{dong2023autoencoders} & 27.06 & 84.66 & 86.1 \\
    Point-BERT~\cite{yu2022point} & 27.09 & 84.11 & 85.6 \\
    Point-MAE~\cite{pang2022masked} & 27.06 & 84.19 & 86.1 \\
    
    \midrule
    \multicolumn{4}{c}{\textit{Parameter efficient fine-tuning (PEFT) with different backbones}}\\
    \midrule    
    \multicolumn{4}{c}{Point-BERT backbone}\\
    \hdashline
    IDPT~\cite{zha2023instance} & 5.69 & 83.50 & 85.3 \\
    DAPT~\cite{zhou2024dynamic} & 5.65 & \textbf{83.83} & \textbf{85.5} \\
    \cgr GFT (\Ours) & \cgr \textbf{3.84} & \cgr 83.41 & \cgr 85.4 \\
    \midrule
    \multicolumn{4}{c}{Point-MAE backbone} \\
    \hdashline
    IDPT~\cite{zha2023instance} & 5.69 & 83.79 & \textbf{85.7} \\
    DAPT~\cite{zhou2024dynamic} & 5.65 & 84.01 & \textbf{85.7} \\
    \cgr GFT (\Ours) & \cgr \textbf{3.84} & \cgr \textbf{84.05} & \cgr 85.6\\
    
    \bottomrule
    \end{tabular}%
  }
  \caption{\textbf{Part segmentation on the ShapeNetPart~\cite{yi2016scalable}}. The mIoU for all classes (Cls.) and instances (Inst.) are reported. \#T-Params
represents the trainable parameters.}
  \label{tab:part_seg}%
\end{table}%
We report the ShapeNetPart~\cite{yi2016scalable} segmentation performance of all of the fine-tuning approaches with mIoU of all classes and instances as in \Cref{tab:part_seg}.
Our method has 3.45 million parameters which is a $\sim$40\% reduction in the parameter size of the baseline methods, establishing itself as the most efficient method.
Despite the smaller model size, GFT competes with the existing methods.

\section{Ablation Studies}
\label{sec:ablation_studies}
\begin{figure}[t]
    \centering
    \includegraphics[width=0.8\linewidth]{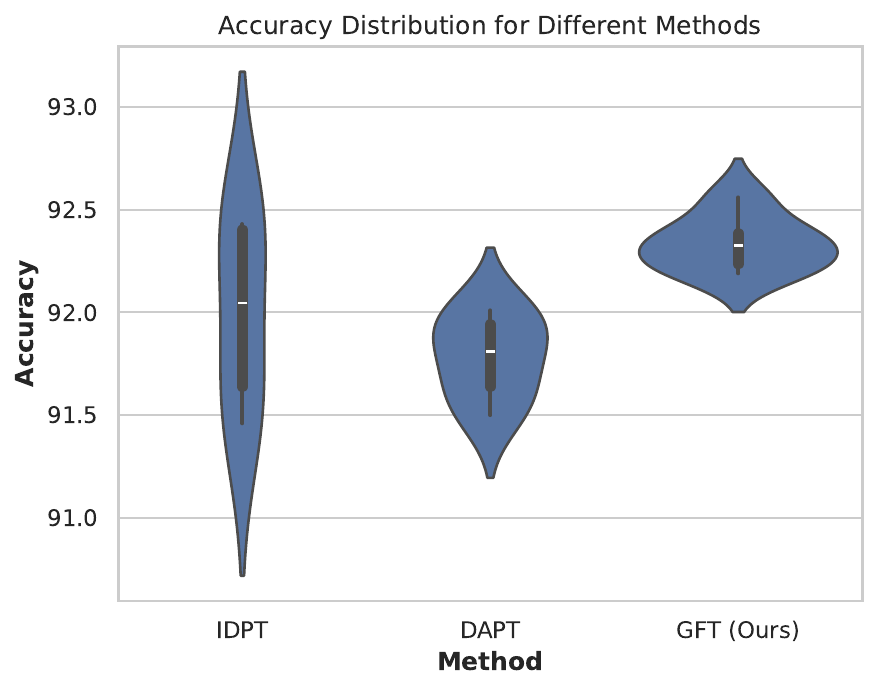}
    \caption{ Violin plot comparing GFT and other PEFT methods across parameter sizes $(0.6M,1.7M)$ range. IDPT and DAPT display higher performance instability as parameters change, whereas GFT demonstrates greater immunity to parameter variation.}
    \label{fig:params_vs_perf}
\end{figure}

For the ablation study, we choose Point-MAE as the backbone, OBJ\_BG as the candidate classification task, and overall accuracy (OA\%) as the evaluation metrics.
Each experiment of \Cref{tab:ablate_components,tab:ablate_x_attention} and of \Cref{fig:params_vs_perf} is subjected to five consecutive seeds within $[0,4]$ for robust comparison.
\underline{Underlined OA} of the tables indicates the final settings of the reported results.
Given the page limits, we have included the additional ablations of the individual components in the supplementary (\Cref{sec:suppl_ablation}).

\paragraph{Analysis of Different Components.}

\begin{table}[t]
  \centering
  \resizebox{\linewidth}{!}{
    \begin{tabular}{ccccc}
    \toprule
    EdgeConv & Cross-Attention & Learnable Prompts & \#T-Params (M) & OA(\%) \\
    \midrule
    \xmark & \xmark & \xmark & 0.30 & 87.26 \\
    \cmark & \xmark & \xmark & 0.48 & 91.67 \\
    \cmark & \cmark & \xmark & 0.62 & 91.81 \\
    \cmark & \xmark & \cmark & 0.59 & 91.94 \\
    \cmark & \cmark & \cmark & 0.73 & \underline{92.56} \\
    \bottomrule
    \end{tabular}%
  }
  \caption{Analysis of the combination of GFT's three different components from \Cref{sec:gft}. Final combination of three gives the best performance.}
  \label{tab:ablate_components}%
\end{table}%

We evaluate the impact of three key components in GFT: EdgeConv for graph feature extraction, cross-attention interaction, and task-specific learnable prompts. The performance of different component combinations is presented in \Cref{tab:ablate_components}.  

Without these components, the model is equivalent to linear probing, where only the task head is fine-tuned, achieving an overall accuracy of 87.26\%. 
Introducing the EdgeConv module significantly improves performance, yielding a 4.41\% gain. 
EdgeConv has a receptive field of $k = 20$, which effectively incorporates the locality features. 
It proves to be highly effective on its own, and the addition of cross-attention interaction and learnable prompts further enhances accuracy. 
Specifically, integrating cross-attention with EdgeConv yields a further 0.14\% improvement, underscoring the benefit of allowing encoder streams to selectively attend to graph features. 
Learnable prompts further enhance performance by 0.27\%, as they offer a greater degree of freedom for graph construction. The final combination of all components achieves a total performance gain of 0.89\% over EdgeConv-only variant, with only a minimal increase of 0.25 million trainable parameters.

\paragraph{Analysis of Interaction Blocks.}

\begin{table}[t]
  \centering
  \resizebox{\linewidth}{!}{
    \begin{tabular}{cccc}
    \toprule
    Layers & Interaction Type & \#T-Params (M) & OA(\%)\\
    \midrule
    $\{1,2,3,4\}$ & Consecutive & 0.73 & 91.98 \\
    $\{9,10,11,12\}$ & Consecutive & 0.73 & 91.64 \\
    $\{1,2,3,..,12\}$ & Consecutive & 1.08 & 92.56 \\
    $\{1,4,7,10\}$ & Sparse & 0.73 & \underline{92.56} \\
    \bottomrule
    \end{tabular}%
  } 
  \caption{Analysis of cross-attention interaction at different layers of Point Transformer: Light sparse interactions achieve similar performance to heavy consecutive interactions.}
  \label{tab:ablate_x_attention}%
\end{table}%

We evaluate the sensitivity of the 12 encoder layers to graph features extracted from the EdgeConv module.
To fine-tune the model, we design two types of cross-attention interactions: consecutive and sparse, as shown in \Cref{tab:ablate_x_attention}. In consecutive interaction, graph features are injected into multiple consecutive layers, whereas in sparse interaction, they are introduced at non-adjacent layers.
\Cref{tab:ablate_x_attention} shows that fine-tuning the first four layers $\{1,2,3,4\}$ improves performance by 0.34\% compared to fine-tuning the last four layers $\{9,10,11,12\}$. 
Since deeper layers in an encoder gradually learns features from coarse to fine, updating the coarse features seems efficacious to adapt new data distributions more effectively.
While fine-tuning all layers may yield the best performance, it is often parameter-inefficient. Instead, sparsely updating a few layers, such as $\{1,4,7,10\}$, achieves similar results.

\paragraph{Parameters vs. Performance.}
An important question to consider is: \textit{What happens to performance when we reduce the parameters of existing PEFT methods to match those of GFT?} As shown in \Cref{fig:params_vs_perf}, scaling down the parameters of DAPT and IDPT leads to a significant drop in overall accuracy (OA), whereas GFT remains more stable.  

Specifically, when reducing the parameter count, IDPT experiences fluctuations of up to 0.97\% in OA, while DAPT fluctuates by 0.51\%. In contrast, GFT exhibits only a 0.37\% variation, indicating greater stability to parameter reductions. Although DAPT’s fluctuation appears closer to GFT, its overall average performance remains lower, reinforcing the efficiency and stability of our method in achieving competitive results with fewer trainable parameters.

\paragraph{Analysis of EdgeConv Features.}
\begin{table}[t]
  \centering
    \begin{tabular}{lccc}
    \toprule
    EdegeConv Dims.  & Depth & \#TP (M) & OA(\%)\\
    \midrule
    $[32,32,32,32]$ & 4 & 0.66 & 92.25 \\
    $[64,64,64]$ & 3 & 0.71 & 91.77 \\
    $[64,64,64,64]$ & 4 & 0.73 & \underline{92.56} \\
    $[64,64,64,64,64]$ & 5 & 0.76 & 91.77 \\
    $[64,128,128,64]$  & 4 & 0.81 & 91.81 \\
    $[128,128,128,128]$  & 4 & 0.92 & 92.19 \\
    $[256,256,256,256]$ & 4 & 1.45 & 92.19  \\
    \bottomrule
    \end{tabular}%
  \caption{Analysis of different dimensions of the EdgeConv feature pyramid. EdgeConv feature pyramid wider than 64 and deeper than 4 leads to overfitting of the fine-tuned tasks. }
  \label{tab:ablate_edgeconv_dims}%
\end{table}%

We investigate the impact of varying the feature dimensions and the depth of the EdgeConv block pyramid, as shown in \Cref{tab:ablate_edgeconv_dims}. Our results indicate that a depth of 4, with each layer having a feature dimension of 64, provides the optimal performance. Increasing the depth or widening the layers does not lead to further improvements, as the model begins to overfit.

\paragraph{Analysis of Cross-attention Dimensions.}
\begin{table}[t]
  \centering
    \begin{tabular}{ccc}
    \toprule
    Cross-attention Dim.  & \#TP (M) & OA(\%)\\
    \midrule
    16 & 0.65 & 91.22 \\
    32  & 0.73 & \underline{92.56} \\
    64  & 0.90 & 92.36 \\
    128 & 1.22 & 92.39 \\
    \bottomrule
    \end{tabular}%
  \caption{Analysis of different dimensions of cross-attention interaction module. Increasing the dimension does not necessarily improve the performance.}
  \label{tab:ablate_xattn_dims}%
\end{table}%
We analyze the impact of varying the dimensions of cross-attention blocks on overall task performance in \Cref{tab:ablate_xattn_dims}.
Increasing the dimension from 16 to 32 results in a 1.34\% improvement in accuracy. 
However, further increasing the dimension beyond 32 does not yield additional performance gains despite the rise in parameter count.
Considering the balance between parameter efficiency and performance, we select 32 as the optimal dimension for our final experimental setup.
\section{Conclusion and Limitation}

In this paper, we introduce Graph Feature Tuning (GFT), a parameter-efficient fine-tuning method that encodes features in the token embedding space and injects them into transformer encoders. 
While our method does not outperform all existing approaches across every task, it remains competitive with state-of-the-art (SOTA) methods while significantly reducing the number of trainable parameters. 
Notably, when the parameters of SOTA methods are reduced to match GFT, they suffer a substantial drop in performance. 
This highlights the effectiveness of our approach, which leverages a combination of smaller components rather than relying on a single, large PEFT module.  

Our findings also indicate the need for a better pretraining dataset than ShapeNet~\cite{girshick2014rich}, which consists of synthetic data. 
While our method, along with other FFTs and PEFTs, performs well on downstream tasks involving synthetic datasets like ModelNet40~\cite{wu20153d}, its performance is less pronounced with real-world object scans from datasets such as ScanObjectNN~\cite{uy2019revisiting}. 
This suggests that improving the pretraining process with a more diverse and representative distribution of real-world scans could further enhance the robustness and generalizability of these methods

Leveraging graph features shows potential for improving segmentation tasks; however, our current work includes only a basic evaluation, leaving room for more in-depth and comprehensive analysis in future research.
Also, our method introduces additional inference FLOPs after fine-tuning.
In the future, we can work with adaptation method similar to LoRA that add zero inference latency.

\section*{Acknowledgement}
Research was sponsored by the Army Research Laboratory and was accomplished under Cooperative Agreement Number W911NF-23-2-0224. The views and conclusions contained in this document are those of the authors and should not be interpreted as representing the official policies, either expressed or implied, of the Army Research Laboratory or the U.S. Government. The U.S. Government is authorized to reproduce and distribute reprints for Government purposes notwithstanding any copyright notation herein.

{
    \small
    \bibliographystyle{ieeenat_fullname}
    \bibliography{main}
}

\clearpage
\setcounter{page}{11}
\setcounter{section}{7}
\maketitlesupplementary

\section{Additional Ablation Study}
\label{sec:suppl_ablation}
Similar to the ablation study in the main paper, we use Point-MAE as the pretrained backbone and OBJ\_BG as the downstream classification task; each experimentation is subjected to five different seed values.

\paragraph{Prompt Length vs. Performance.}

\begin{table}[t]
  \centering
    \begin{tabular}{ccc}
    \toprule
    Learnable Prompt Length  & \#TP (M) & OA(\%)\\
    \midrule
    10 & 0.72 & 91.91 \\
    25  & 0.72 & 91.74 \\
    50  & 0.73 & \underline{92.56} \\
    100 & 0.75 & 92.05 \\
    \bottomrule
    \end{tabular}%
  \caption{Analysis of prompt length for GFT analysis. Prompt length beyond 50 hurts the performance.}
  \label{tab:ablate_prompt_len}%
\end{table}%

As shown in \Cref{tab:ablate_prompt_len}, we evaluate the sensitivity of our method to prompt length by experimenting with four different values: $\{10, 25, 50, 100\}$. The optimal performance is achieved with a prompt length of 50, which we use for our reported results. Increasing the prompt length does not necessarily yield better performance. Unlike patch tokens, these prompts lack locality information but offer greater degrees of freedom in graph construction. However, an excessive number of prompts may introduce unnecessary edges, which could negatively impact downstream tasks.

\paragraph{EdgeConv KNN}

\begin{table}[t]
  \centering
    \setlength{\tabcolsep}{1mm}
  \fontsize{9}{11}\selectfont
    \begin{tabular}{cccc}
    \toprule
    \multirow{2}{*}{EdegeConv KNN}  & \multicolumn{2}{c}{Training Complexity} & \multirow{2}{*}{OA(\%)} \\
    \cmidrule{2-3}
    & Epoch Time (secs) & Memory (GB) \\
    \midrule
     5 & 15.5 & 6.9 & 91.77 \\
     10 & 16.5 & 7.1 & 91.67 \\
     20 & 18.0 & 8.7 & \underline{92.56} \\
     40 & 23.5 & 10.4 & 92.22 \\
    \bottomrule
    \end{tabular}%
  \caption{Analysis of different dimensions of the EdgeConv feature pyramid. Increasing the neighbours in KNN imposes  higher training time and memory complexities.}
  \label{tab:supp_ablate_edgeconv_knn}%
\end{table}%

The performance of GFT is also influenced by the number of neighbors used for message passing.
Changing the number of neighbors does not affect the trainable parameters but changes the time and computational complexities.
To analyze this, we conduct an ablation study with different values of $k \in \{5,10,20,40\}$ in KNN, as shown in \Cref{tab:supp_ablate_edgeconv_knn}.  
Given 50 prompt tokens and 128 patch tokens, we have a total of 178 tokens and message passing with the 20 nearest neighbors yields the best results.  
While increasing the number of neighbors raises both time and memory complexity, it does not lead to further improvements in performance. 

\paragraph{Tokens Projections for Task Head}
\begin{table}[t]
  \centering
    \setlength{\tabcolsep}{1mm}
  \fontsize{9}{11}\selectfont
    \begin{tabular}{ccccc}
    \toprule
    \multicolumn{3}{c}{Feature Pooling Techniques} &  \multirow{2}{*}{\#TP (M)} & \multirow{2}{*}{OA(\%)} \\
    \cmidrule{1-3}
    $T_{cls}$ & Patch Pooling & Prompt Pooling \\
    \midrule
    \cmark & & & 0.54 & 90.29 \\
     
     \cmark & \cmark & & 0.64 & 91.98 \\
     \cmark & \cmark & \cmark & 0.73 & \underline{92.56} \\
    \bottomrule
    \end{tabular}%
  \caption{Analysis of token projections for classification task. The combinationion of all of the feature pooling gives the best result.}
  \label{tab:supp_ablate_proj}%
\end{table}%
We explore combinations of three options for projecting features from Point Transformers for classification: \( T_{cls} \), patch pooling, and prompt pooling. 
Previous fine-tuning methods primarily relied on \( T_{cls} \) and patch pooling as projections for the task head~\cite{pang2022masked,zhou2024dynamic,zha2023instance,dong2023autoencoders,yu2022point}. 
However, GFT has learnable prompts where task-specific features are accumulated. 
We achieve the best results by leveraging all three projection options, as shown in \Cref{tab:supp_ablate_proj}.

\section{Training Details}
\label{sec:supp_train_details}

\begin{table*}[t]
    \centering
        \begin{tabular}{lcccc}
            \toprule
            \multirow{2}{*}{Configuration} & \multicolumn{3}{c}{Classification} & \multirow{2}{*}{Segmentation (ShapeNetPart)} \\
            \cmidrule{2-4}
            & ScanObjectNN & ModelNet & ModelNet Few-shot & - \\
            \midrule
            Optimizer & AdamW & AdamW & AdamW & AdamW \\
            Learning rate (LR) & 5e-4 & 5e-4 & 1e-3 & 2e-4  \\
            LR scheduler & Cosine & Cosine & Cosine & Cosine \\
            Warmup LR & 1e-6 & 1e-6 & 1e-6 & 1e-6 \\ 
            Warmup epochs & 10  & 10  & 10  & 10  \\
            Minimum LR & 1e-6 & 1e-6 & 1e-6 & 1e-6 \\
            Weight decay & 5e-2 & 5e-2 &  1e-4 & 5e-2 \\
            Training epochs & 300 & 300 & 150 & 300 \\
            Batch size & 32 & 32 & 32 & 32 \\
            \midrule
            Number of points & 2048 & 1024 & 1024 & 2048 \\
            Number of patches/tokens & 128 & 64 & 64 & 128 \\
            Point patch size & 32 & 32 & 32 & 32 \\
            \midrule
            \multicolumn{5}{c}{\textit{GFT Configurations}}\\
            \midrule
            Learnable prompt length & 50 & 50 & 50 & 50 \\
            EdgeConv KNN size & 20 & 20 & 20 & 20 \\
            EdgeConv feat dims. & $[64,64,64,64]$ & $[64,64,64,64]$ & $[64,64,64,64]$ & $[64,64,64,64]$ \\
            EdgeConv FFN dim & 256 & 256 & 256 & 256 \\
            Cross-attention dim & 32 & 32 & 32 & 32 \\
            Cross-attention heads & 2 & 2 & 1 & 2 \\
            \bottomrule
        \end{tabular}%
    \caption{Training details for the fine-tuning tasks.}
    \label{tab:supp_train_details}
\end{table*}
\Cref{tab:supp_train_details} provides the comprehensive training details for our method across different classification and segmentation tasks. 
Classification models have been optimized using cross-entropy loss, while segmentation models utilize negative log-likelihood loss. We follow a similar training approach to earlier PEFT methods~\cite{zhou2024dynamic,zha2023instance}.

\section{Architecture for Segmentation Head}
\label{sec:suppl_seg_head}
\begin{figure}[t]
    \centering
    \includegraphics[width=1.0\linewidth]{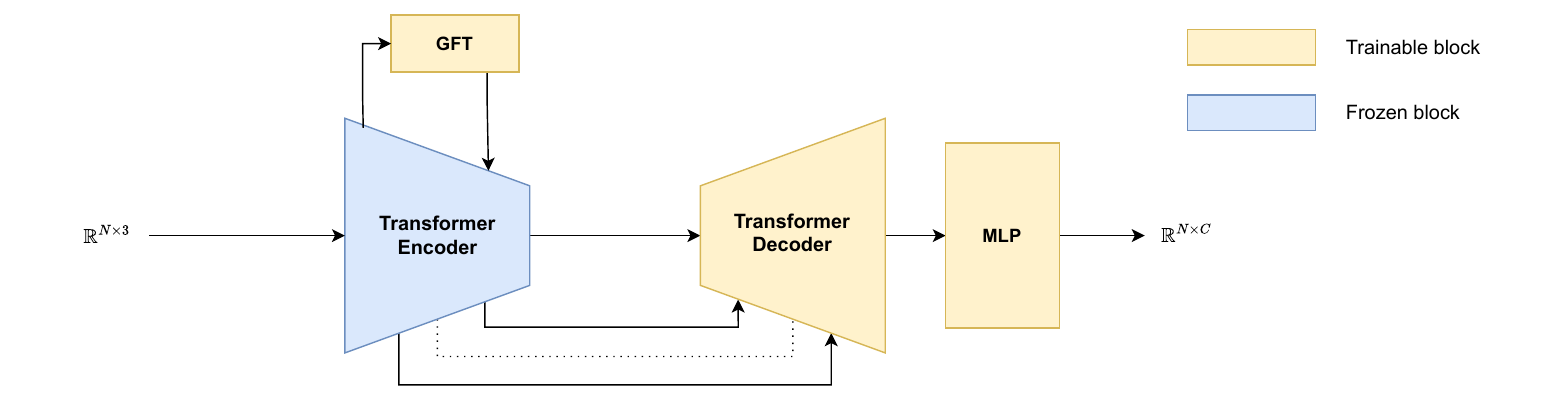}
    \caption{\textbf{Overall architecture of the segmentation model.} The segmentation head utilizes intermediate features from the encoder layers, processes them through transformer decoders, and finally passes them to an MLP layer to predict the class labels.}
    \label{fig:decoder_for_segmentation}
\end{figure}

Our part segmentation task head consists of two main components: (1) a transformer decoder, composed of stacked self-attention blocks, which receives input from the encoder representations through skip connections and down-projection layers; and (2) a multi-layer perceptron (MLP) that outputs class predictions for each point, producing a tensor of shape $\mathbb{R}^{N \times C}$, where $N$ is the number of points and $C$ is the number of classes as given in \Cref{fig:decoder_for_segmentation}.

\section{Vizualization of Attention Maps}
\label{sec:suppl_vizualizations}
\begin{figure*}[h]
    \centering
    \includegraphics[width=1\linewidth]{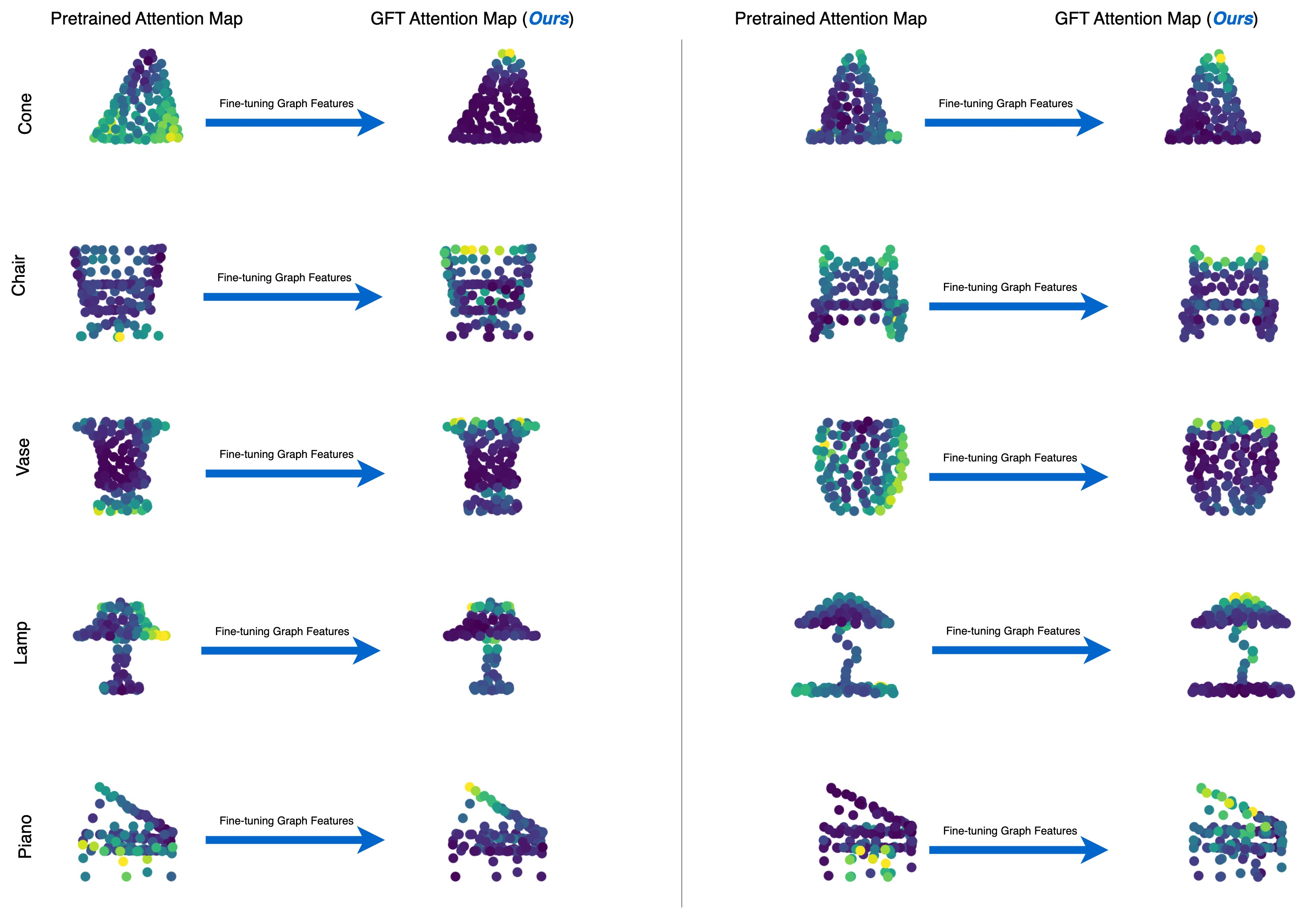}
    \caption{Fine-tuning leads to a shift in attention, redirecting focus from arbitrary regions to task-relevant, discriminative parts of the objects—such as the backrest for chairs, apex for cones, mouth for vases, and fallboard for pianos. In contrast, the pretrained model exhibits uncertainty, with attention scattered across different parts.}
    \label{fig:ablate_attention_maps}
\end{figure*}

As shown in the \Cref{fig:ablate_attention_maps}, during fine-tuning of the pretrained backbones, we analyze the attention maps from patch tokens to the CLS token ($T_{cls}$). When the pretrained weights are kept frozen, fine-tuning the task-level graph features guides the attention to consistently focus on discriminative regions. For example, attention maps for cones highlight the apex, while those for chairs emphasize the backrest. In contrast, the pretrained model often struggles to identify a consistent region of interest—sometimes focusing on one part of the object, and other times on another—indicating a lack of clear task-specific understanding.

\end{document}